\documentclass{article}

\usepackage{microtype}
\usepackage{graphicx}
\usepackage{subfigure}
\usepackage{booktabs} % for professional tables

\usepackage{hyperref}

\usepackage[accepted]{article}

\icmltitlerunning{Self-Supervised Encoder for Fault Prediction in Electrochemical Cells}

\usepackage{mathptmx}

\usepackage[american]{babel}

\begin{document}

\twocolumn[
\icmltitle{Self-Supervised Encoder for Fault Prediction in Electrochemical Cells}



\icmlsetsymbol{equal}{*}

\begin{icmlauthorlist}
\icmlauthor{Daniel Buades Marcos}{poly,r2}
\icmlauthor{Soumaya Yacout}{poly}
\icmlauthor{Said Berriah}{r2}
\end{icmlauthorlist}

\icmlaffiliation{poly}{Polytechnique Montréal, 2900 Édouard Montpetit Blvd, Montréal, H3T 1J4, QC, Canada.}
\icmlaffiliation{r2}{R2 Inc., 380 Saint-Antoine St W Suite 7500, Montréal, H2Y 3X7, QC, Canada}

\icmlcorrespondingauthor{Daniel Buades Marcos}{\href{mailto:daniel@buad.es}{daniel@buad.es}}

\icmlkeywords{Machine Learning}

\vskip 0.3in
]



\printAffiliationsAndNotice{}  % leave blank if no need to mention equal contribution

\begin{abstract}
Predicting faults before they occur helps to avoid potential safety hazards.
Furthermore, planning the required maintenance actions in advance reduces operation costs.
In this article, the focus is on electrochemical cells.
In order to predict a cell's fault, the typical approach is to estimate the expected voltage that a healthy cell would present and compare it with the cell's measured voltage in real-time.
This approach is possible because, when a fault is about to happen, the cell's measured voltage differs from the one expected for the same operating conditions.
However, estimating the expected voltage is challenging, as the voltage of a healthy cell is also affected by its degradation -- an unknown parameter.
Expert-defined parametric models are currently used for this estimation task.
Instead, we propose the use of a neural network model based on an encoder-decoder architecture.
The network receives the operating conditions as input.
The encoder's task is to find a faithful representation of the cell's degradation and to pass it to the decoder, which in turn predicts the expected cell's voltage.
As no labeled degradation data is given to the network, we consider our approach to be a self-supervised encoder.
Results show that we were able to predict the voltage of multiple cells while diminishing the prediction error that was obtained by the parametric models by 53\%.
This improvement enabled our network to predict a fault 31 hours before it happened, a 64\% increase in reaction time compared to the parametric model.
Moreover, the output of the encoder can be plotted, adding interpretability to the neural network model.
\end{abstract}

\section{Problem Description}
\label{Problem Description}

Electrolysis is the process of decomposing a chemical product into various byproducts by applying an electrical current \cite{Wendt1999}.
It takes place in \emph{electrolyzers}, which are systems composed of multiple \emph{electrochemical cells}.
These cells act similarly to resistors: electrical current passing through them causes a voltage drop.
The magnitude of this voltage drop depends on the operating conditions and the cell's degradation, increasing steadily when a fault is about to happen.

Faults in electrochemical cells may become safety hazards.
In order to diminish their occurrence, they are usually replaced every four years.
This heuristic comes from the statistical analysis of the average lifetime of past cells, which represents an aggregation of data from multiple cells.
Thus, it does not take into consideration the specificity of each cell, which is needed to take action concerning their maintenance.
In order to implement a more efficient strategy that adequately considers such specificity, the cell's degradation must be monitored.
However, degradation can only be determined directly by performing offline, and sometimes destructive, tests \cite{Causserand2010}.
Our objective is to use non-invasive methods that do not require the full stoppage of the electrolyzer.
As such, the cell's degradation must be inferred indirectly from other measurable properties.
In this article, our approach relies on predicting the voltage that a healthy cell would present ($\widehat{V_t}$) and comparing it to the cell's measured voltage in real-time ($V_t$).
If there is a divergence between the two of them, a fault is signaled.
The divergence threshold and the type of fault diagnosed depend on the shape of the divergence, following a set of rules proposed by experts.

In order to predict the voltage of a healthy cell, the following limitations are considered:

\begin{enumerate}

	\item The cell's voltage drifts slowly over time due to its degradation.

	\item Even for the same operating conditions and degradation level, the voltage of a cell differs from the ones of similar cells.
		This difference is induced by disparities in manufacturing, installation procedures, and other factors that cannot be directly quantified.
		The latter is referred to in this article as the specificity of each cell.
					
	\item There is a delay between a change in the operating conditions and the response of the cell.
		In order to account for this delay, the operating conditions at the previous time-steps must be considered by the prediction model.

	\item Inside the electrolyzer, cells are electrically connected in series and they all receive the same chemical input.
		This means that the only measurable variable that is different for each cell is their voltage.
		This voltage is stored according to the position of each cell in the electrolyzer, and not according to the unique identification number of each cell.
		Moreover, as previously stated, cells' degradation data is not available.

	\item In order to predict faults, the predicted voltage must be independent from the measured voltage.
		This limitation entails that the measured voltage cannot be used as an input to the prediction model.

	\item Electrolyzers are shut down and restarted many times throughout their lifetime.
		During each shutdown, the data collection is stopped.
		Any cell may be substituted or exchanged at the discretion of the plant's operator, without being reflected in the data.
		The intervals when the electrolyzer is operating are called cycles and, as all its cells are under tension, no changes are performed by the operator.

\end{enumerate}

\section{Previous Work}
\label{Previous Work}

To be able to find the right model and techniques that apply to our problem, we first need to understand the system we are working with, which is a chemical electrolyzer.
As previously stated, chemical electrolyzers are composed of multiple cells, where the electrolysis process takes place.
There are different cell technologies, the most efficient being \emph{Ion-Exchange Membranes}, on which our article is focused.
In these cells, the anode and the cathode are separated by a semi-permeable membrane that does not permit both electrolytes to mix, yet allows the ions needed for the electrolysis to travel across \cite{Paidar2016}.
As time passes, holes start to appear in the membrane, and the electrodes start to lose their coating, thus increasing the voltage of the cell \cite{Jalali2009}.
As the degradation advances, it reaches a point where an undesirable reaction between the solution in the anode and the solution in the cathode starts to occur.
This reaction is characterized by a spike in the cell's voltage \cite{TheChlorineInstitute2018}.
It is a dangerous situation requiring the electrolyzer to be stopped immediately.

Therefore, detecting anomalies in the cell's voltage leads to predicting faults in the cell.
As previously stated, we do so by predicting the expected voltage that a healthy cell would present and then comparing it with its actual measured voltage.
The better the accuracy of the model used for predicting such voltage, the earlier it is possible to signal the fault.

In order to calculate the expected cell's voltage ($\widehat{V_t}$), the underlying chemical reaction needs to be approximated.
This function is composed of many parameters, which need to be estimated.
Some of them are operation-specific, i.e., identical for all the cells that perform the electrolysis for the same operating conditions.
However, as per limitation 2, there are also cell-specific parameters.
Moreover, limitation 1 implies that there are also degradation-specific parameters that change over time as the cell degrades.

In order to define a model that accounts for all these parameters, three types of data are needed: \emph{operation data}, \emph{cell-specific data}, and \emph{degradation-specific data}.
Nonetheless, as per limitations 4 and 5, we neither have cell-specific data nor degradation-specific data that we can use.
Thus, if only operation-specific parameters are used, the same voltage would be predicted for each cell.
This is not an acceptable result, because the specificity of each cell would be lost.
In order to overcome this problem, the current approach relies on fitting a different model for each cell and retraining it periodically in order to update the degradation-specific parameters.
As per limitation 6, the retraining frequency must be at least once per cycle.

An expert-defined parametric model is currently used for this task \cite{Tremblay2012}.
Operation-specific parameters are defined by the experts, while cell and degradation-specific parameters are estimated at the beginning of each cycle by using a linear regression.
This model has the advantage of being simple, thus needing fewer observations to train than equivalent non-parametric models \cite{Veillette2010}.
This is crucial, as the observations used at each cycle for training the model cannot be used for predicting the cells' voltage.
Therefore, during the training period, it is not possible to detect faults either.
This model also has easy to explain results.
Nonetheless, it requires an understanding of the chemical process to define it.
Suppositions in the model entail an accuracy loss, as they do not account for all the details present in a production environment.

An alternative approach is to use non-parametric Machine Learning (ML) techniques.
Support Vector Machines (SVM) were used to predict the voltage of a \emph{chlor-alkali} cell and explore its response to different operating conditions \cite{Kaveh2009}.
They obtained better accuracy than parametric models, but the scope of their work was limited to a single cell in a controlled lab environment with pre-defined operating conditions.
They carried out a similar study using artificial Neural Networks (NNs) \cite{Kaveh2008}.
However, their network only had two hidden layers and a reduced number of neurons, falling behind recently developed networks. 

NNs are non-parametric models composed of multiple mathematical entities called \emph{neurons}.
Their name comes from the fact that they are inspired by how biological brains work \cite{Goodfellow2016}.
Neurons are grouped in layers.
In the most basic neural architecture, called the Multi-Layer Perceptron (MLP), each neuron of a particular layer is connected to all the neurons of the previous layer.
A scalar, called \emph{weight}, quantify each connection.
The neuron's output is the sum of the values of each neuron from the previous layer multiplied by their respective weights.
Up to this point, the output of the model would be a linear combination.
In order to approximate non-linear functions, the output of each neuron is transformed by an activation function \cite{Nielsen2015}.
Indeed, when using non-linear activation functions and enough neurons combined with data, NNs are considered to be universal approximators \cite{Pinkus1999}.
For our application, this is primordial: we are no longer forced to make assumptions about the underlying chemical function.
There are many architectures derived from the general MLP, each of them specifically tailored to work with a different kind of input data and objective.
Examples of these architectures are Recurrent Neural Networks (RNNs) and neural encoders, which are both used in this article.

As per limitation 3, we are interested in RNNs for our application.
An RNN is a particular neural architecture that deals with sequential data.
It accomplishes so by keeping an internal state that is updated for every time-step of the sequence \cite{Karpathy2015}.
This internal state serves the purpose of a memory, allowing past information to persist in the network.
In order to predict the output of a certain time-step, it considers the input for that time-step and the internal state from the previous time-steps.
However, RNNs struggle when dealing with long temporal sequences \cite{Bengio1994}.
Long-Short Term Memory (LSTM) networks are a type of RNNs that does not suffer from this problem \cite{Hochreiter1997}.
LSTMs are capable of modifying their internal state, either by removing or by adding new information, through the use of learnable gates \cite{Olah2015}.
This flexibility makes LSTMs to be more commonly used in practice than RNNs.

A neural encoder is a type of neural architecture whose objective is to take an input vector and reduce its dimensionality to a desired one.
It is usually paired with a decoder.
The decoder receives the output of the encoder and transforms it to minimize an objective function.
The whole network is trained \emph{backpropagating} the loss of the decoder \cite{Rumelhart1985}.
If the objective of the decoder is to reconstruct the original input vector, it is called an autoencoder \mbox{\cite{Hinton2006}}.
A review of different types of autoencoders is presented in the work of Tschannen et al. \cite{Tschannen2018}.
Autoencoders are often used for anomaly detection \cite{Sakurada2014} and noise reduction \cite{Vincent2008}.
They are also applicable to time sequences in combination with LSTMs \cite{Malhotra2016}.

\subsection{Originality}
\label{Originality}

In this article, we develop a method to apply NNs to the voltage prediction of all the electrolyzer's cells.
The model works in a production environment, where operating conditions are not pre-defined, and each cell has a different level of degradation.
The model also has better accuracy than the parametric model currently in use.
Moreover, it does not need more observations per cycle for starting to make predictions.

As NNs require more data than parametric models to approximate a function, we take a different approach.
Instead of fitting a different model for each cell and cycle, we propose a neural network that is trained with data currently available, thus avoiding retraining once deployed.
This network is based on an \emph{encoder-decoder} architecture, where we substitute the \emph{decoder} by a \emph{predictor} -- a subnetwork that predicts the cell's voltage.
Therefore, instead of training the network by minimizing the reconstruction error of the input sequence, we train it by minimizing the error between the voltage predicted by the network and the measured voltage.

In order to overcome all the six limitations previously mentioned, the originality of our approach in comparison with the already existing approaches is:

\begin{enumerate}

	\item The encoder subnetwork addresses limitations 1, 2, and 6.
		It does so by finding two features that represent the specificity of each cell and that are updated at each cycle to account for the degradation.

	\item The predictor addresses limitation 3, as it takes into account the temporality of the observations.

	\item Together, the encoder and the predictor address limitations 4 and 5.
		The predictor does not use the voltage as an input, yet it is still able to predict a different voltage for each cell, despite using the same operating conditions as input.
		It accomplishes so by taking the output of the encoder as an additional input, which is unique for each cell.
		Hence, the voltage prediction is not biased by the cell's measured voltage.

\end{enumerate}

\section{Data Preparation}
\label{Data Preparation}

\subsection{Features}
\label{Features}

Data is collected from two different sources.
The first source is the plant's control system, which registers three features common to all the cells in the electrolyzer.
These features are the electrical current that passes through the cells ($I$), and the temperature plus the concentration of the caustic at the outlet of the electrolyzer ($T$ and $X$ respectively).
\linebreak The second source is the output of sensors that measure the voltage of each cell individually ($V$). 

Cells' measurements are carried out sequentially, which means that the controller reads a sensor and then proceeds to read the next one.
This process is repeated for every cell in the electrolyzer and takes between one and two seconds to loop over all the cells.
The plant's controller, on the other hand, does not follow a strict pattern of data collection.
Each sensor connected to it has a different sampling rate, varying from one to around thirty seconds, which results in data observations that are misaligned and sampled at different intervals.
In order to solve this problem, we downsample and align the measurements to the minute.
This approach provides enough resolution to detect possible faults, as faults develop in the scale of hours.
It also reduces the amount of unnecessary training data and facilitates the deployment of the model in slower computing processors, as the latency requirements are less strict.

The resulting data after alignment is tabulated.
Each row represents a different time of observation, and there is a different column for each electrolyzer's feature and cell.
An electrolyzer is usually formed of around 160 cells.
An excerpt of this data is presented in Table~\ref{table: Excerpt of the dataset}.

\begin{table}[t]
		\caption{Excerpt of the dataset where \emph{n} represents the last observation, and \emph{m} the last electrolyzer's cell.}
		\label{table: Excerpt of the dataset}
		\vskip 0.1in
		\begin{center}
		\begin{small}

		\begin{tabular}{lcccccc}
		\toprule
		Index & $I$ & $T$ & $X$ & $V_1$ & \dots & $V_m$ \\
		\midrule
		
		1 & 1.163 & 76.668 & 31.949 & 2.475 & \dots & 2.468 \\
		2 & 1.887 & 76.555 & 31.945 & 2.532	& \dots	& 2.518 \\
		3 & 2.072 & 76.501 & 31.941 & 2.562 & \dots & 2.549 \\
		4 & 2.036 & 76.449 & 31.937 & 2.560 & \dots & 2.554 \\
		5 & 2.425 & 76.397 & 31.937 & 2.577 & \dots & 2.562 \\
		\dots & \dots & \dots & \dots & \dots & \dots & \dots \\
		$n$-4 & 16.311 & 88.742 & 32.672 & 3.380 & \dots & 3.318 \\
		$n$-3 & 16.310 & 88.733 & 32.673 & 3.380 & \dots & 3.318 \\
		$n$-2 & 16.310 & 88.724 & 32.673 & 3.380 & \dots & 3.318 \\
		$n$-1 & 16.309 & 88.715 & 32.674 & 3.381 & \dots & 3.317 \\
		$n$ & 16.309 & 88.706 & 32.675 & 3.381 & \dots & 3.317 \\
		
		\bottomrule
		\end{tabular}

		\end{small}
		\end{center}
		\vskip -0.2in
\end{table}

\subsection{Cycles}
\label{Cycles}

\begin{figure}[t]
		\caption{Evolution of the electrical current of the electrolyzer during a complete cycle.
		The startup phase is plotted in blue and the operation phase, in orange.
		Notice the length's difference between both phases.}
		\label{fig: Evolution of the electrical current}

		\vskip 0.2in
		\begin{center}
		\centerline{\includegraphics[width=\columnwidth]{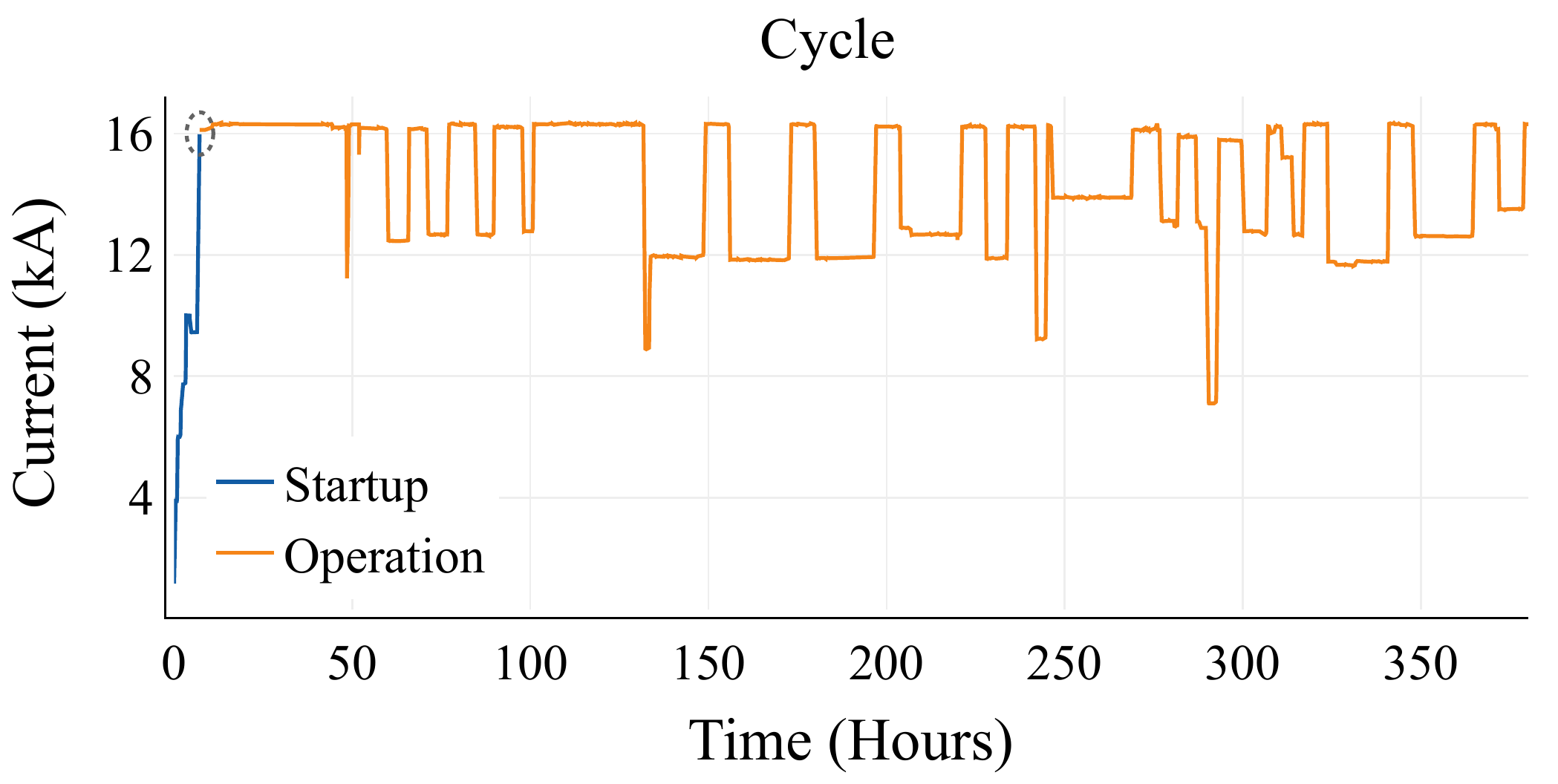}}

		\end{center}
		\vskip -0.2in
\end{figure}

Electrolyzers are stopped and started many times during their lifetime due to production constraints, changing demand, work shifts, or maintenance requirements.
The interval of time between a consecutive start and stop is known as a \emph{cycle}, and its length ranges from some hours to several weeks, depending on those conditions.
Each cycle is divided into two phases of different lengths -- the \emph{startup} and the \emph{operation} phase.
Both cycle's phases are shown in Figure~\ref{fig: Evolution of the electrical current}.

\begin{figure}[t]
		\caption{Voltage during the startup phase of the same cell (Y) for three different cycles (A, B, C).}
		\label{fig: Voltage startup same cell}

		\vskip 0.2in
		\begin{center}
		\centerline{\includegraphics[width=\columnwidth]{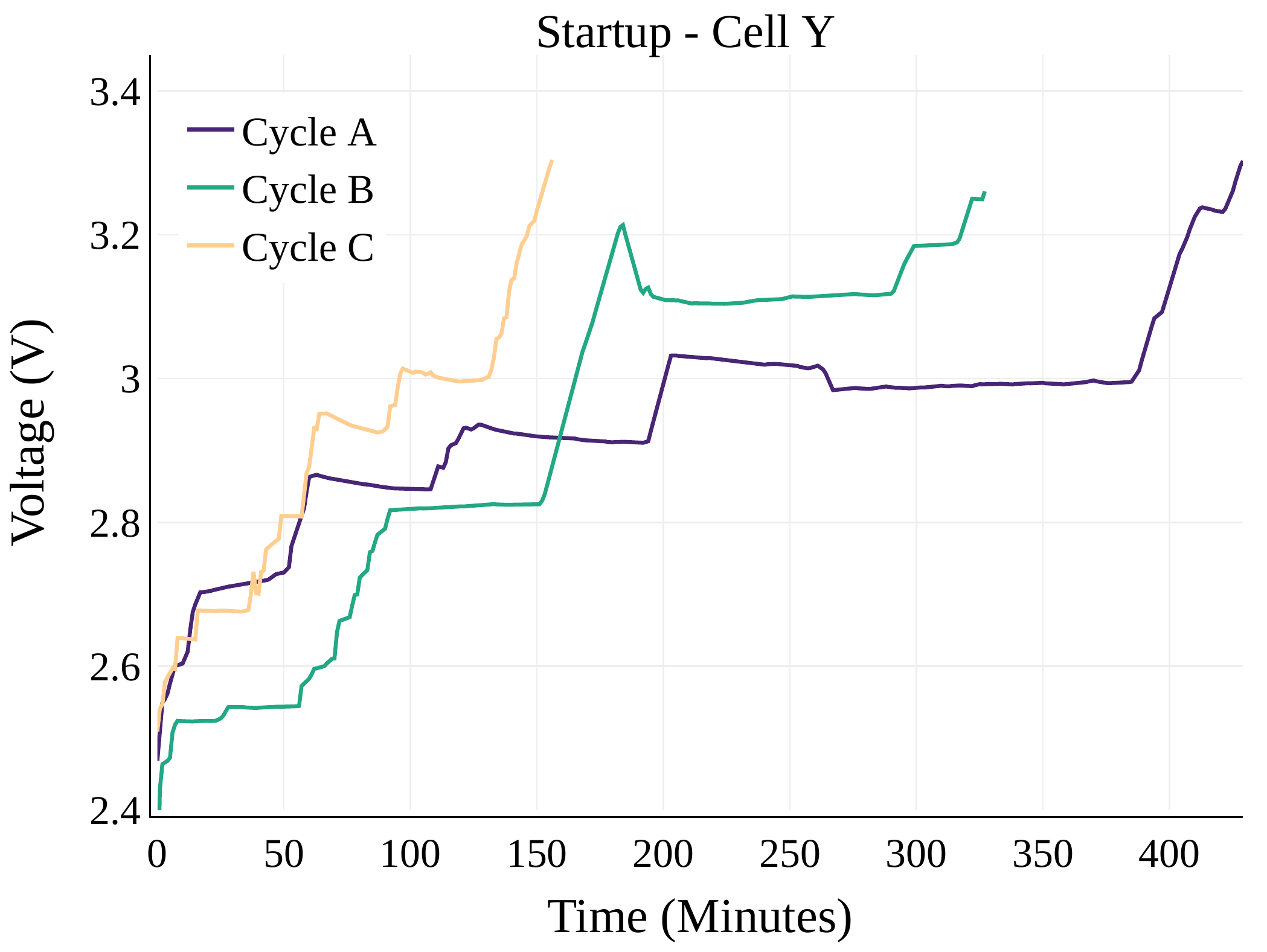}}

		\end{center}
		\vskip -0.2in
\end{figure}

\begin{figure}[t]
		\caption{Voltage during the startup phase of three different cells (X, Y, Z) for the same cycle (A).}
		\label{fig: Voltage startup same cycle}

		\vskip 0.2in
		\begin{center}
		\centerline{\includegraphics[width=\columnwidth]{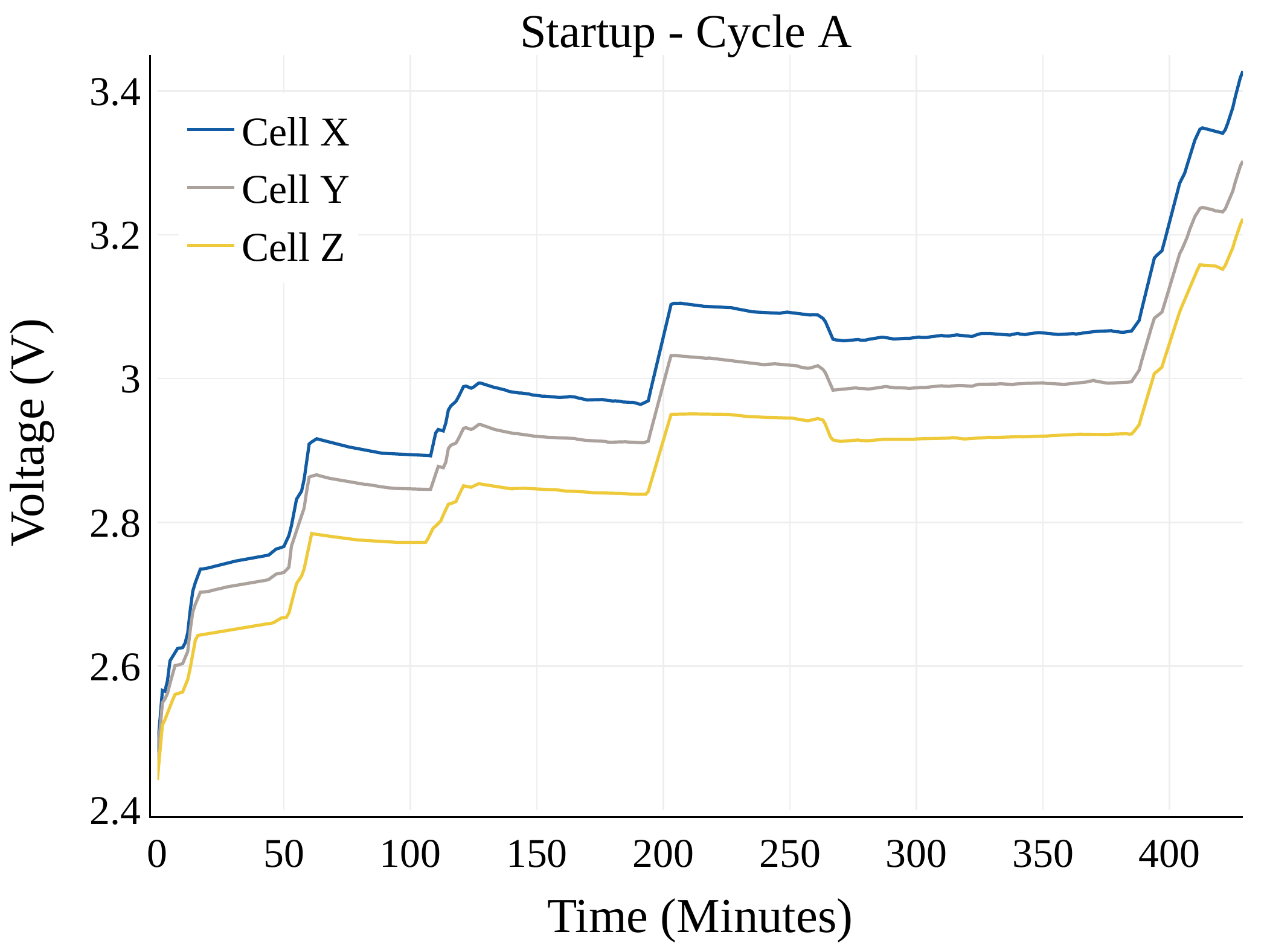}}

		\end{center}
		\vskip -0.2in
\end{figure}

The startup phase occurs when the electrical current increases from zero to 16 kA.
The 16 kA threshold, defined by an expert, represents the moment when the cells reach their full production conditions.
The rate at which the current increases differs for each startup, due to changes in the operating practices decided by the plant's operators.
The length of this phase ranges anywhere from 20 minutes to 12 hours, which complicates making comparisons between different startups.
This length's difference is shown in Figure~\ref{fig: Voltage startup same cell}.
As previously mentioned, each cell responds differently to the same operating conditions, resulting in a different voltage increase during the startup.
The voltage increases of three different cells are shown in Figure~\ref{fig: Voltage startup same cycle}.

Note that cycles A, B, and C, as well as cells X, Y, and Z, are used as examples through the article.
We chose these cycles because their operating conditions are significantly different.
The three cells are chosen because they have different levels of degradation.
Please note that their color schemes remain the same for all the figures in the article.

The operation phase includes the rest of the cycle.
The electrical current varies between 7 kA and 16 kA.
We are especially interested in predicting the voltage during this phase, as a cell's fault would cause a significant disturbance.

\section{Methodology}
\label{Methodology}

\subsection{Data Processing}
\label{Data Processing}

As previously stated, when the electrolyzer is shut down, data collection is stopped, and so, it appears as missing data in the file.
A new cycle is detected if the time difference between two consecutive observations is longer than 10 minutes and the following experts' conditions are met:

\begin{enumerate}
		\item The electrical current reaches 16 kA during the cycle.
		\item The duration of the startup phase is less than 12 hours.
		\item The operation phase has at least the same duration as the startup phase.
\end{enumerate}

Data files are processed in order to ensure the satisfaction of these conditions, following the procedure outlined in Algorithm~\ref{alg: Detect Possible Cycles} and Algorithm~\ref{alg: Validate Cycles}.

\begin{algorithm}[b]
	
	\caption{Detect Possible Cycles}
	\label{alg: Detect Possible Cycles}
	\begin{algorithmic}

		\STATE {\bfseries Input:} $time\_array$ with $n$ observations
		\STATE {\bfseries Initialize:} $initial\_cycle\_index \gets 0$, $i \gets 0$
		\STATE {\bfseries Initialize:} $cycle\_list \gets [\ ]$

		\WHILE{i $<$ lenght($time\_array$)}

			\STATE $diff_{i} \gets time\_array_{(i+1)} - time\_array_{(i)}$

			\IF {$diff_{i} > 10$ minutes}

				\STATE $cycle \gets time\_array[initial\_cycle\_index:i]$
				\STATE $cycle\_list$.append($cycle$)
				\STATE $initial\_cycle\_index \gets i+1$

			\ENDIF
			
			\STATE $i \gets i + 1$
		\ENDWHILE

	\end{algorithmic}
\end{algorithm}

\begin{algorithm}[b]
		\caption{Validate Cycles}
		\label{alg: Validate Cycles}
		\begin{algorithmic}

		\STATE {\bfseries Input:} $cycle\_list$

		\FOR{each $cycle$ in $cycle\_list$}

				\IF {any observation in $cycle$ has $current > 16\ kA$}
						\STATE $idx \gets$ First observation where $current > 16\ kA$

						\IF {$idx$ $\leq$ 12 hours}
								\STATE $startup \gets cycle[0:idx]$
								\STATE $operation \gets cycle[idx:-1]$

								\IF {lenght($operation$) $\geq$ length($startup$)}
										\STATE $cycle$ is a valid cycle

								\ENDIF
						\ENDIF
				\ENDIF
		\ENDFOR

		\end{algorithmic}
\end{algorithm}

Once the data is structured in cycles, with their startup and operation phases defined, we proceed to scale it.
For each data column, we use the \emph{unity-based normalization} scaling method, also known as \emph{Min-Max Scaling}.
It scales the data linearly, so all the values are in the range [0,1] \cite{Raghav2018}.
We do not use more sophisticated scaling techniques since our data is not normally distributed and since outliers are already removed in the original database.
The column is scaled as \begin{small}$X_{scaled}=(X-min)/(max-min)$\end{small}.

The minimum (\emph{min}) and maximum (\emph{max}) values use for scaling each feature are common for all the production plants and are determined by experts.
They are the same for all the electrolyzers, so it is possible to use the same trained model for all of them.
Missing observations appear as \emph{`NaN'} and are substituted by the value of `-1' in order to keep them outside of the scaler's range.
After following this procedure, a subset of the resulting data is shown in Table~\ref{table: subset of processed data from cycle A}
The final step is to export the data to TFRecord files, which is a binary format developed by Google and optimized for the preprocessing tf.data.Dataset API of TensorFlow 2.0 \cite{Google2019}.
We export a different file for each cycle, containing all the cell's observations.

\begin{table}[t]
		\caption{Subset of data from Cycle A after being processed.
				Data has been scaled and missing observations, replaced by `-1'.
				Note that, as expected, the input features at each time-step are the same for cells \emph{one} and \emph{m}, 
				where \emph{m} represents the last electrolyzer's cell.}
		\label{table: subset of processed data from cycle A}
		\vskip 0.1in
		\begin{small}
		\begin{center}

		\begin{tabular}{lcccccc}
		\toprule
		Cell & Phase & Index & $I$ & $T$ & $X$ & $V$ \\
		\midrule
		
		1 & $Startup$ & 1 & 0.010 & 0.417 & 0.790 & 0.441 \\
		 &  & 2 & 0.054 & 0.413 & 0.790 & 0.450 \\
		 &  & 3 & 0.066 & 0.411 & 0.789 & 0.455 \\
		 &  & 4 & 0.064 & -1 & 0.789 & 0.454 \\
		 &  & \dots & \dots & \dots & \dots & \dots \\
		 &  & $i^a$ & 0.920 & 0.790 & 0.851 & 0.587 \\ \cmidrule{2-7}
		 
		 & $Oper.$ & 1 & 0.928 & 0.806 & 0.852 & 0.587 \\
		 &  & 2 & 0.928 & 0.811 & -1 & 0.587 \\
		 &  & 3 & 0.928 & 0.815 & 0.852 & 0.587 \\
		 &  & 4 & 0.928 & 0.819 & 0.852 & 0.586 \\
		 &  & \dots & \dots & \dots & \dots & \dots \\
		 &  & $j^b$ & 0.940 & 0.881 & 0.856 & 0.586 \\

		\midrule
		\dots & \dots & \dots & \dots & \dots & \dots & \dots \\
		\midrule

		$m$ & $Startup$ & 1 & 0.010 & 0.417 & 0.790 & 0.440 \\
		 &  & 2 & 0.054 & 0.413 & 0.790 & 0.448 \\
		 &  & 3 & 0.066 & 0.411 & 0.789 & 0.453 \\
		 &  & 4 & 0.064 & -1 & 0.789 & 0.454 \\
		 &  & \dots & \dots & \dots & \dots & \dots \\
		 &  & $i^a$ & 0.920 & 0.790 & 0.851 & 0.573 \\ \cmidrule{2-7}

		 & $Oper.$ & 1 & 0.928 & 0.806 & 0.852 & 0.575 \\
		 &  & 2 & 0.928 & 0.811 & -1 & 0.575 \\
		 &  & 3 & 0.928 & 0.815 & 0.852 & 0.574 \\
		 &  & 4 & 0.928 & 0.819 & 0.852 & 0.574 \\
		 &  & \dots & \dots & \dots & \dots & \dots \\
		 &  & $j^b$ & 0.940 & 0.881 & 0.856 & 0.576 \\
		
		\bottomrule
		\end{tabular}
		\end{center}
		
		% Footnotes
		\textsuperscript{$a$} $i$ : Last \emph{startup} observation of cycle A \\
		\textsuperscript{$b$} $j$ : Last \emph{operation} observation of cycle A

		\end{small}
		\vskip -0.2in
\end{table}

\subsection{Model Architecture}
\label{Model Architecture}

As previously explained, we want to predict the voltage over the whole operation phase of the cycle, so the training must only be performed during the startup phase.
However, neural networks are complex non-linear methods that require a large amount of data to converge to an optimal solution.
Hence, only using the startup data for training a new model at each cycle does not provide satisfactory results.

We have collected data from multiple cells across multiple cycles of different electrolyzers.
Ideally, we would like to use all this data to train our network, but it is not straightforward to do so.
The reason is because, as previously explained, each cell presents a different response to operating conditions, but no feature or labeled data is available to differentiate them.

One possible approach is to train a base naïve model and then use transfer learning to retrain the last few layers of the model.
However, we would need to do that for each cell at the beginning of each cycle, which is a computationally intensive task that would complicate the deployment \cite{Weiss2016}.
Instead, we propose a different approach inspired by how an expert may look at many cells and differentiate them based solely on the evolution of their voltage during the startup sequence.
Based on this idea, we deduce that it is possible to use the voltage of each cell during the startup phase to infer meaningful features that characterize such cells.

These features account for both the specificity of the cell and its degradation and are used as input for the predictor model.
This model also takes the operating conditions in order to predict the cell's voltage during the operation phase.
The cell's voltage is only used as input during the startup phase for inferring the features.
Indeed, the predictor model does not use the cell's voltage during the operation phase, so the predictions are not biased.
Given enough data belonging to multiple cells and cycles, we would cover the whole subspace of possible features.
Thus, new cells presented to the model would behave like cells that the model had already seen before, and so, it could also predict their voltage accurately.

By using this approach, it is possible to train a single model that works for all the cells.
Such a model predicts a different voltage for each cell based not only on the operating conditions but also on the specific cell's features inferred during the startup.
This effectively avoids the need to retrain the model for each cycle and cell.
At the same time, it allows us to train it beforehand with the whole dataset that we already have.

In order to implement this model, we propose a neural network formed by two subnetworks: a self-supervised encoder and a decoder -- or voltage predictor.
These two subnetworks are developed in detail in the two two following subsections.
The algorithm required for training such a network is developed in Section~\ref{section: Training Procedure}.

\subsubsection{Encoder}
\label{section: Encoder}

The encoder subnetwork is the key part of this model.
Its goal is to infer the features that characterize the behavior of a particular cell during a certain cycle, using only its startup phase.
It is a self-supervised method, as it does not need labeled degradation data.
This subnetwork does not have a loss function on its own.
Instead, it is trained with the loss backpropagated from the voltage predictor subnetwork.

At its core, it is a form of performing a dimensionality reduction.
However, the encoded vector -- the dimensionality-reduced vector -- is not the result of a statistical procedure, but the optimal representation that eases the learning of the predictor subnetwork.

We use the three input features from the control plant -- temperature, caustic concentration, and electrical current -- as well as the voltage of the cell.
These inputs are needed to standardize the cell's voltage to the specific operating conditions of each startup. All in all, the shape of the input vector is [720 time-steps, 4 features].
The masking layer forces the successive layers to ignore a time-step if all the features of that time-step are equal to a masked value, which is `-1' in order to filter the time-steps added during batch padding.
In order to account for the temporality of the sequence, the next layer is a Long Short-Term Memory (LSTM).
After it, two dense layers are chained, smoothing the transition to the final two-positional encoded result.

\begin{figure}[t]
		\caption{Encoder subnetwork}
		\label{fig: Encoder subnetwork}

		\vskip 0.2in
		\begin{center}
		\centerline{\includegraphics[width=\columnwidth]{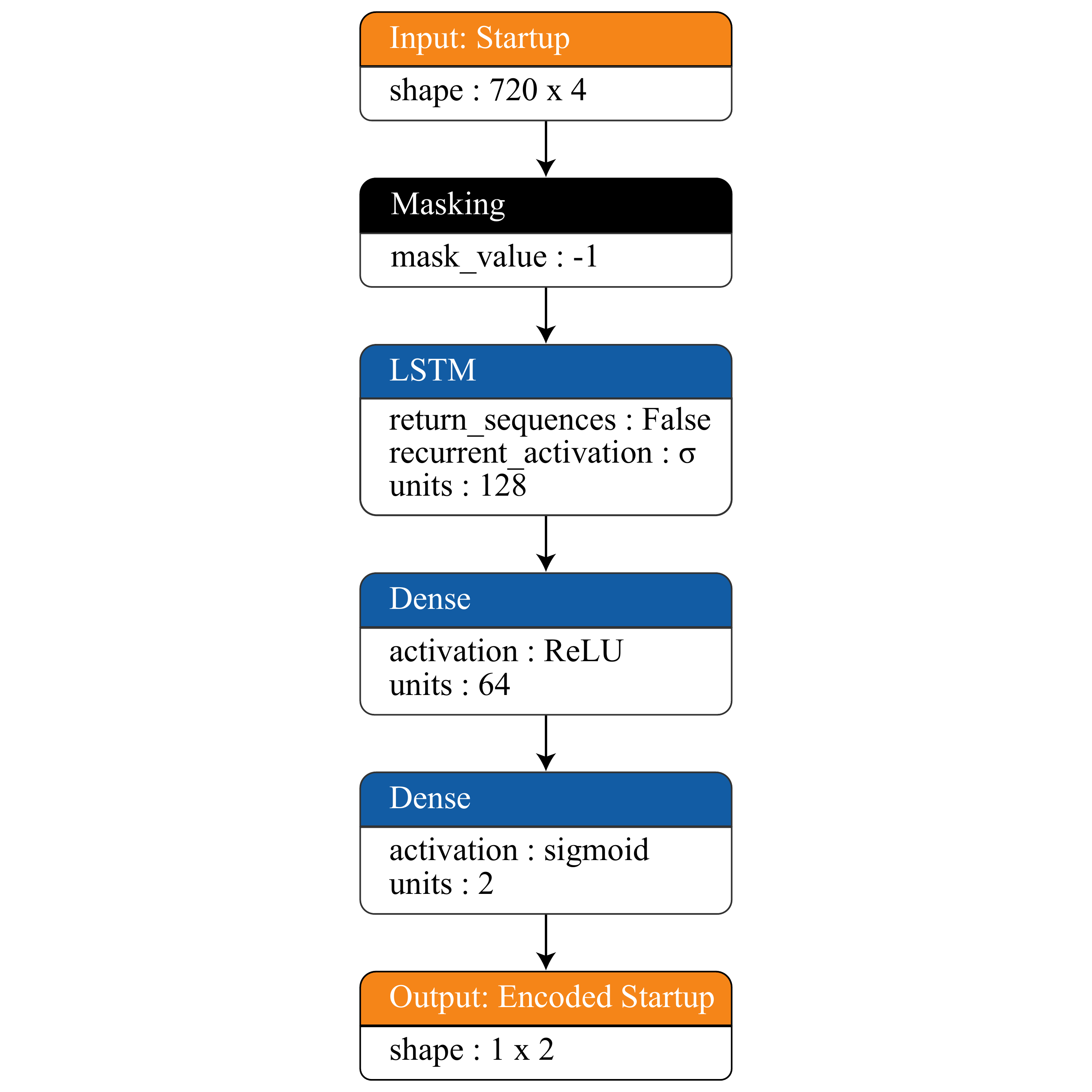}}
		\end{center}
		\vskip -0.2in
\end{figure}

Its output is a vector of coordinates [X, Y] for each cell and startup, with a shape of (1 time-step, 2 features).
Moreover, these coordinates can be represented in a graph, providing an insight into the decision process taken by the network.
The characteristics of the layers that compose this subnetwork -- in TensorFlow's terminology -- are depicted in Figure~\ref{fig: Encoder subnetwork}.

\subsubsection{Predictor}
\label{section: Predictor}

The predictor subnetwork, as its name indicates, is responsible for predicting the cell's voltage.
It takes two inputs: a window of time-steps from the operation phase and the encoded representation of the cell's startup phase.
We have heuristically determined that a window of four observations is enough to represent the dynamics of the chemical phenomena behind the cell's response.

The encoded cell's startup is repeated four times and concatenated with the window of operation features.
Note that the voltage is not included in the window of operation features.
Two LSTM layers are used to find temporal correlations between the observations of each window.
Two dense layers follow, in order to output the predicted voltage.
The output layer has a sigmoid activation function, as the output voltage was scaled previously to the range [0, 1].

\begin{figure}[t]
		\caption{Predictor subnetwork}
		\label{fig: Predictor subnetwork}

		\vskip 0.2in
		\begin{center}
		\centerline{\includegraphics[width=\columnwidth]{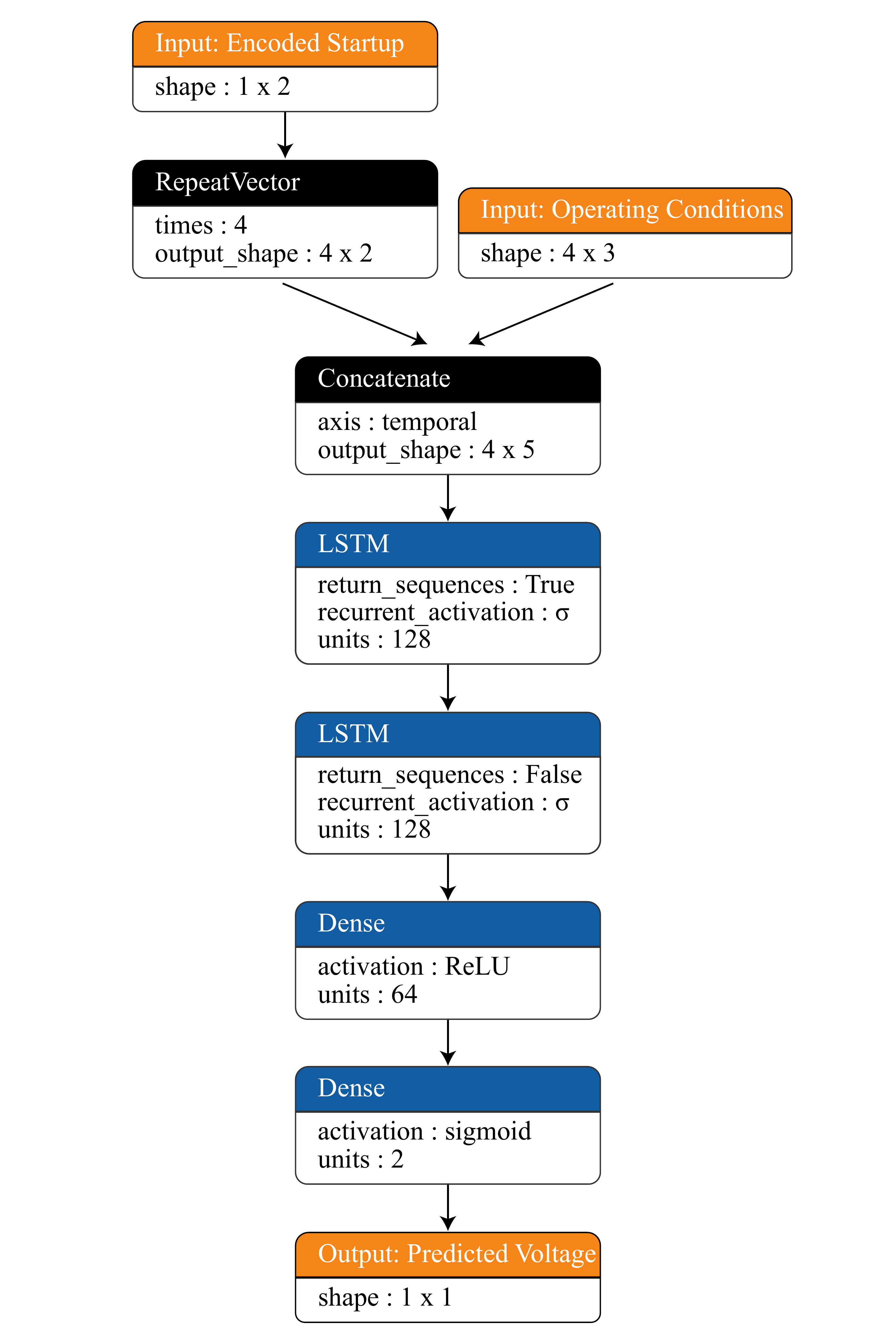}}
		\end{center}
		\vskip -0.2in
\end{figure}

The whole model is trained by minimizing the loss between the voltage predicted by this subnetwork and the measured voltage.
The Adam optimizer \cite{Kingma2014} and the backpropagation algorithm are used.
For this subnetwork to get a good accuracy in the voltage prediction, the encoder must learn a faithful representation of the characterization of the cell.
Figure~\ref{fig: Predictor subnetwork} presents its layers' parameters.

\subsection{Training Procedure}
\label{section: Training Procedure}

We start with the data presented in Table~\ref{table: subset of processed data from cycle A}.
As previously mentioned in Section~\ref{section: Predictor}, in order to account for the temporality of the chemical reaction, we first need to group four consecutive time-steps into a single data entry.
This data entry is known as \emph{window} or \emph{observation}.
In order to train a model with this kind of data, we define a custom training loop.
Algorithm~\ref{alg: Stochastic Training Algorithm} presents the training loop for stochastic training -- one observation per forward-backward pass.

However, such a training loop is not efficient.
In order to leverage the parallel computation made possible by GPUs, we use \emph{mini-batch} training instead.
In this mode of training, many observations are grouped in a \emph{batch} and ingested by the GPU at the same time.\linebreak

% Dirty trick to align the algorithm header with the table
% https://tex.stackexchange.com/questions/321765/how-to-increase-algorithms-top-space-to-page-margin
\makeatletter
\newcommand\fs@spaceruled{\def\@fs@cfont{\bfseries}\let\@fs@capt\floatc@ruled
  \def\@fs@pre{\vspace{1\baselineskip}\hrule height.8pt depth0pt \kern2pt}%
  \def\@fs@post{\kern2pt\hrule\relax}%
  \def\@fs@mid{\kern2pt\hrule\kern2pt}%
  \let\@fs@iftopcapt\iftrue}
\makeatother
\floatstyle{spaceruled}% Select new float style
\restylefloat{algorithm}

\begin{algorithm}[t]
	\caption{Stochastic Training Algorithm}
	\label{alg: Stochastic Training Algorithm}
	\begin{algorithmic}

	\STATE {\bfseries Input:} Directory with $n$ cycles in TFRecord format

	\FOR {each $file$ in folder}

		\STATE $cycle$ $\gets$ TFRecord.load($file$)

		\FOR {each $cell$ in $cycle$}
			\STATE $cell\_startup$, $cell\_operation$ $\gets$ $cell$.split()

			\FOR {each window in split$(cell\_operation)$}
				\STATE $operating\_conditions \gets window[!voltage]$
				\STATE $cell\_voltage \gets window[voltage][-1]$\vspace{0.5em}

				\COMMENT{Forward pass}

				\STATE $encoded\_startup \gets$ encoder$(cell\_startup)$

				\STATE $predicted\_voltage \gets$ predictor$($
				\STATE \hspace{4em} $operating\_conditions,$
				\STATE \hspace{4em} $encoded\_startup)$

				\STATE $loss \gets$ mean\_squared\_error$($
				\STATE \hspace{4em} $predicted\_voltage,$
				\STATE \hspace{4em} $cell\_voltage)$ \vspace{0.5em}

				\COMMENT{Backward pass}

				\STATE $update\_network\_weights(loss)$

			\ENDFOR
		\ENDFOR
	\ENDFOR

	\end{algorithmic}

\end{algorithm}

In order to make it possible to tweak different parameters effortlessly, we decided to generate the windows and batches in real-time during training.
The procedure is similar to that followed when doing \emph{data augmentation}, which is used extensively in computer vision \cite{Taylor2017}.
While the GPU is processing a batch of windows, the CPU is already processing the next batch.
The GPU is fast, so the CPU quickly becomes the bottleneck in such a pipeline.
In order to overcome this issue, we use the {tf.data.Dataset} API to parallelize the input pipeline.
We trained the network with two GPUs in parallel and a combined batch size of 1024 observations.
Thanks to this parallelized CPU input pipeline, the utilization of both GPUs was consistently over 90\%.

Three operations are crucial to make the network converge efficiently: \emph{padding}, \emph{shuffling}, and \emph{window striding}.

\subsubsection{Padding}
\label{section: Padding}

Not every cycle's startup has the same duration.
However, all the batches that we feed to the GPU must have the same number of time-steps.
We solve this problem by padding the sequences -- adding `-1' values at the end of each observation's corresponding startup.
This way, all the startups have the same duration of 720 minutes, which is the maximum duration of a startup.
This padding value is later ignored by the masking layer of the encoder subnetwork, so it does not affect the results.
\linebreak Table~\ref{table: Padded Startup} shows the startup phase of cycle A after being padded.

\begin{table}[t]
		\caption{Padded startup. $i$ represents the last startup observation before padding.}
		\label{table: Padded Startup}
		\begin{center}
		\begin{small}

		\begin{tabular}{lcccc}
		\toprule
		Index & $I$ & $T$ & $X$ & $V$ \\
		\midrule
		
		1 & 0.010 & 0.417 & 0.790 & 0.441 \\
		2 & 0.054 & 0.413 & 0.790 & 0.450 \\
		3 & 0.066 & 0.411 & 0.789 & 0.455 \\
		4 & 0.064 & -1.000 & 0.789 & 0.454 \\
		\dots & \dots & \dots & \dots & \dots \\
		i & 0.920 & 0.790 & 0.851 & 0.587 \\
		i+1 & -1 & -1 & -1 & -1 \\
		... & \dots & \dots & \dots & \dots \\
		719 & -1 & -1 & -1 & -1 \\
		720 & -1 & -1 & -1 & -1 \\

		\bottomrule
		\end{tabular}

		\end{small}
		\end{center}
		\vskip -0.2in
\end{table}

\subsubsection{Shuffling}
\label{section: Shuffling}

We observed that the time required by the network to converge to an optimal solution was reduced considerably by the introduction of shuffling in the training process.
There is an explanation for this behavior.
Without shuffling, most batches only have observations from a single cell and cycle of the same electrolyzer.
This situation constitutes a problem because the gradient updates at each batch are very different.
However, after introducing shuffling, each batch has now observations from different cells, cycles, and electrolyzers.
As a result, we obtain less noise in the backpropagated gradient and a smoother training loss, which helps the network weights to converge.

However, due to the considerable number of observations in the training dataset, it is not possible to perform the shuffling operation entirely in-memory.
In order to overcome this limitation, we combine two strategies:

\begin{enumerate}
		\item \underline{Shuffle Buffer}: instead of shuffling the whole dataset, we only shuffle one subset of observations at a time.
				This subset is the shuffle buffer, and its number of observations is limited by the size of the computer's RAM.
				After the observations have been shuffled, we give them to the network for training.
				Once the buffer is depleted, a new subset is read, and the same operation is performed again.

		\item \underline{Interleaving}: as previously stated, we have a different file for each cycle and electrolyzer.
				We randomly chose a file, read an observation from it, and add it to the shuffle buffer.
				This process is repeated in a loop.
				Thanks to using interleaving, the shuffle buffer is filled with observations from multiple cycles and electrolyzers, thus increasing the randomness of the batches given to the network.
\end{enumerate}

\subsubsection{Window Striding}
\label{section: Window Striding}

As previously shown in Algorithm~\ref{alg: Stochastic Training Algorithm}, the encoder subnetwork updates its weights for each batch of operation observations.
However, the encoder task of inferring the cell's features during the startup is more complex than that of the predictor.
Hence, it is more efficient to train the network with fewer observations per cycle and more different startup sequences.
In order to solve this, we increase the stride of our windowing function.
The \emph{stride} is the number that defines how many windows of the sequence are ignored between two consecutive training observations.
For example, a stride of 64 means that, for each cell and cycle, we only take one window every 64.

\subsection{Testing Procedure}
\label{section: Testing Procedure}

\subsubsection{Parametric Model}
\label{section: Parametric Model}

We use the parametric model as the baseline for comparing the results obtained by our neural network.
This model predicts the cell's voltage ($\hat{V}$) following the equation presented in Table~\ref{eq: EDE model}.
The advantage of this model over other parametric models is the fewer required observations for getting an equivalent accuracy.
This is crucial since its training data is limited to the startup phase of each cycle.

\begin{table}[h]
		\caption{Parametric model}
		\label{eq: EDE model}
		\vskip 0.15in
		\begin{center}
		\begin{small}
			\begin{sc}

				$\hat{V}=u_0+[k+(90-T)\ast C_t+(32-X)\ast C_x\ ]*I/A$
				\vskip 0.15in
		
				\begin{tabular}{c}
				\midrule
				Parameters \\
				\midrule
				\end{tabular}
				\end{sc}

		\begin{tabular}{l}
		$C_t$ : Temperature correction factor [$V/^{\circ}C * m^2/kA$] \\
		$C_x$ : Caustic correction factor [$V/\% * m^2/kA$] \\
		$u_0$ : Cell's equilibrium voltage [$V$] \\
		$k$   : Load dependent resistance [$V * m^2/kA$] \\	
		$A$   : Membrane's surface area [$m^2$] \\
		\end{tabular}

		\vskip 0.15in
		\begin{sc}
		\begin{tabular}{c}
		\midrule
		Inputs \\
		\midrule
		\end{tabular}
		\end{sc}

		\begin{tabular}{l}
		$I$ : Electrical Current [$kA$] \\
		$T$ : Temperature [$^{\circ}C$] \\
		$X$ : Concentration [\%] \\
		\end{tabular}
		
		\end{small}
		\end{center}
\end{table}

The cell's manufacturer gives parameter \emph{$A$}, which for the cells in our dataset is 2.721.
Parameters \emph{$C_t$} and \emph{$C_x$} are estimated for each plant by experts.
For our test plant, \emph{$C_t$} = 0.0016 and \emph{$C_x$} = -0.0031.
Parameters $u_0$ and $k$ depend on the degradation of each specific cell and must be estimated at the beginning of each cycle.
In order to estimate them, a linear regression is fitted by minimizing the sum of least squares: \begin{small}$\sum{(\hat{V}-V)}^2$\end{small} with all of the cycle's startup observations.

\subsubsection{Method}
\label{section: Method}

All the tests were carried out on data collected from three years of an electrolyzer.
During this period, the electrolyzer went through 40 different cycles.
As the electrolyzer has 160 cells, there are 6400 combinations of cells and cycles in the dataset.
The data used follows the structure previously presented in Table~\ref{table: subset of processed data from cycle A}.

\begin{figure}[t]
		\caption{Prediction Flowchart}
		\label{fig: Prediction Flowchart}

		\vskip 0.2in
		\begin{center}
		\centerline{\includegraphics[width=\columnwidth]{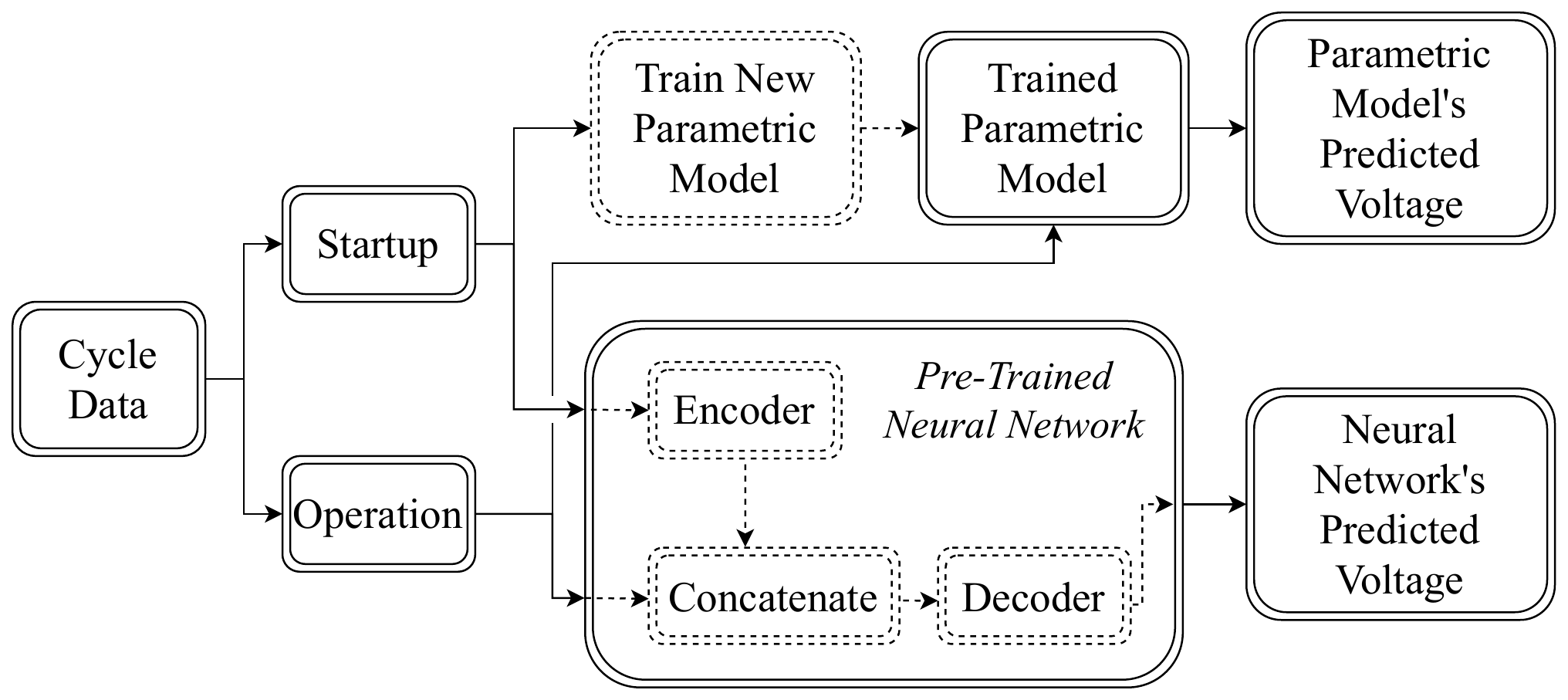}}

		\end{center}
		\vskip -0.2in
\end{figure}

A new parametric model was fitted for each cell during the startup phase of each cycle, following Equation 2.
Consequently, this means that 6400 different models were trained.
In contrast, with our approach, only one neural network model was trained.
The network was trained with data from six different electrolyzers from the same plant, each one having different cells and startup sequences.
In total, around 43200 combinations of cells and cycles were seen by the network during training.
Of course, the electrolyzer used for testing was neither used for training nor for validation to decide when to stop training the network.
For making the predictions, the network receives the startup sequence, but no retraining is performed.
Figure~\ref{fig: Prediction Flowchart} illustrates the differences between the process followed by the parametric model and the one followed by the neural network.

\section{Results and Discussion}
\label{section: Results and Discussion}

\subsection{Network Insight}
\label{section: Network Insight}

\begin{figure}[t]
		\caption{Evolution of the encoding of cells X, Y, and Z along cycles A, B, and C.
				The axes of the graph do not have units, as they are the result of a dimensionality reduction.
				They take values in the interval [0,1] since the encoder's output is a sigmoid function.}
		\label{fig: Startup Encoding}

		\vskip 0.2in
		\begin{center}
		\centerline{\includegraphics[width=\columnwidth]{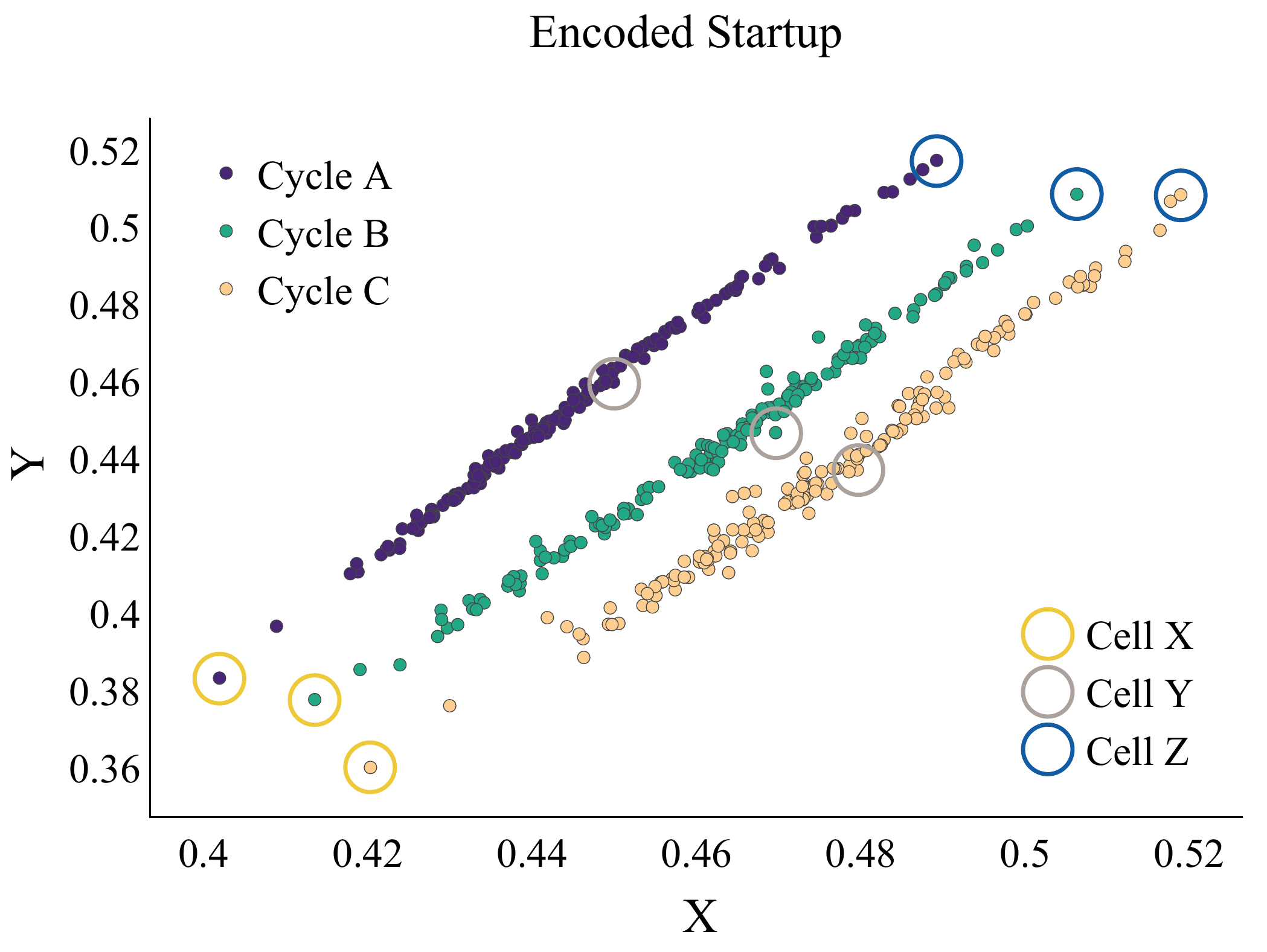}}

		\end{center}
		\vskip -0.2in
\end{figure}

We start by discussing the encoder's results.
In Figure~\ref{fig: Startup Encoding}, it is possible to see the evolution of the encoding of cells X, Y, and Z along cycles A, B, and C of the testing electrolyzer.
Each point presented in Figure 8 corresponds to the encoded startup sequence of a specific cell and cycle.
As previously explained, each cell has its own specificity and degradation.
Let us take the example of cell Z to showcase the veracity of the encoder's results.
We know thanks to an expert that cell Z is the most degraded cell of the whole electrolyzer for these three cycles.
In the graph, we can clearly see this, as cell Z is plotted at the cycle's extreme.
Since the startup's sequences of the testing electrolyzer are different from those of the training electrolyzers, we prove that the encoder is effectively learning the degradation of the cells and not only memorizing the startup sequences.

Furthermore, the relative position of a cell compared to the rest of the cells of that cycle stays stable over time.
This circumstance can be used as a safeguard for the prediction.
Indeed, if we notice that the predicted voltage of a cell presents a significant error, we can check in the graph if the relative position of that cell has changed between the last cycle and the current one.
If there is a significant variation, the voltage prediction error is most likely due to a problem with the encoder subnetwork and not with the cell.
This is why we say that the network's results can be interpreted.

\begin{figure}[t]
		\caption{Predicted voltage for cells X, Y, and Z during a period of one week belonging to cycle A.
				The solid lines represent the real voltage of each cell, while the dotted lines represent the predicted voltage by the network.}
		\label{fig: Predicted Voltage}

		\vskip 0.2in
		\begin{center}
		\centerline{\includegraphics[width=\columnwidth]{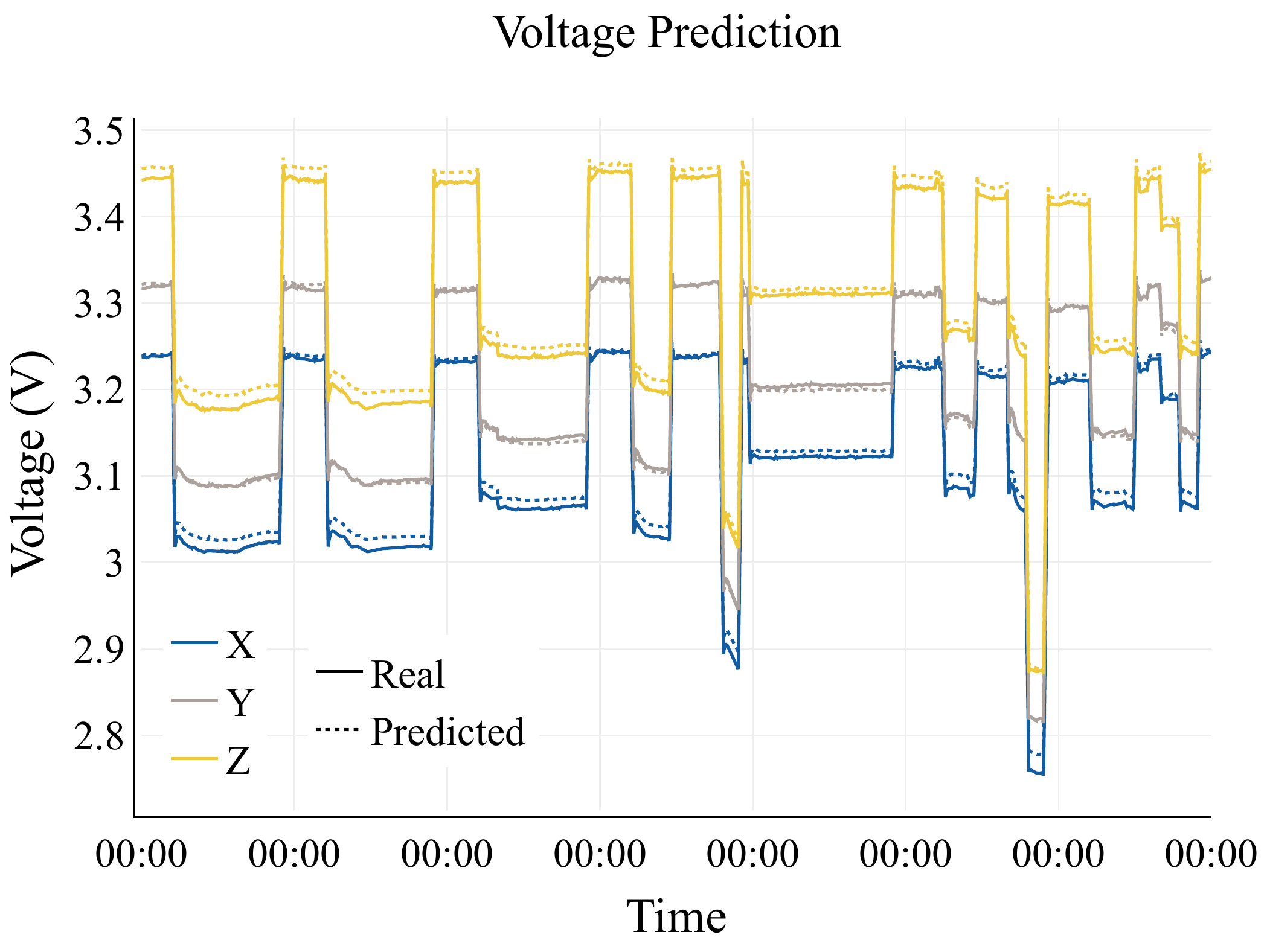}}

		\end{center}
		\vskip -0.2in
\end{figure}

In Figure~\ref{fig: Predicted Voltage}, we present the output voltage of the three cells highlighted in Figure~\ref{fig: Startup Encoding} during a subset of the operation phase of cycle A.
We show that our model is capable of predicting their respective voltages.
We can also see that the cell X, whose startup was plotted at the top of the cycle in Figure~\ref{fig: Startup Encoding}, present the lowest voltage of the three.
Cell Y was encoded in the middle in Figure~\ref{fig: Startup Encoding} and, indeed, its voltage is between the two other cells.
Similarly, cell Z was at the bottom of the cycle in Figure~\ref{fig: Startup Encoding}, and it presents the highest voltage in Figure~\ref{fig: Predicted Voltage}.
This result is consistent in our tests with different cells.
Thus, we can conclude that the magnitude of the predicted voltage is related to the position learned by the encoder.

\begin{figure}[t]
		\caption{Extract of cycle A.
				The upper subgraph shows the absolute error between the predicted and the measured voltage, averaged over Cells X, Y, and Z.
				The lower subgraph presents the electrical current.}
		\label{fig: Models vs Electrical Current}

		\vskip 0.2in
		\begin{center}
		\centerline{\includegraphics[width=\columnwidth]{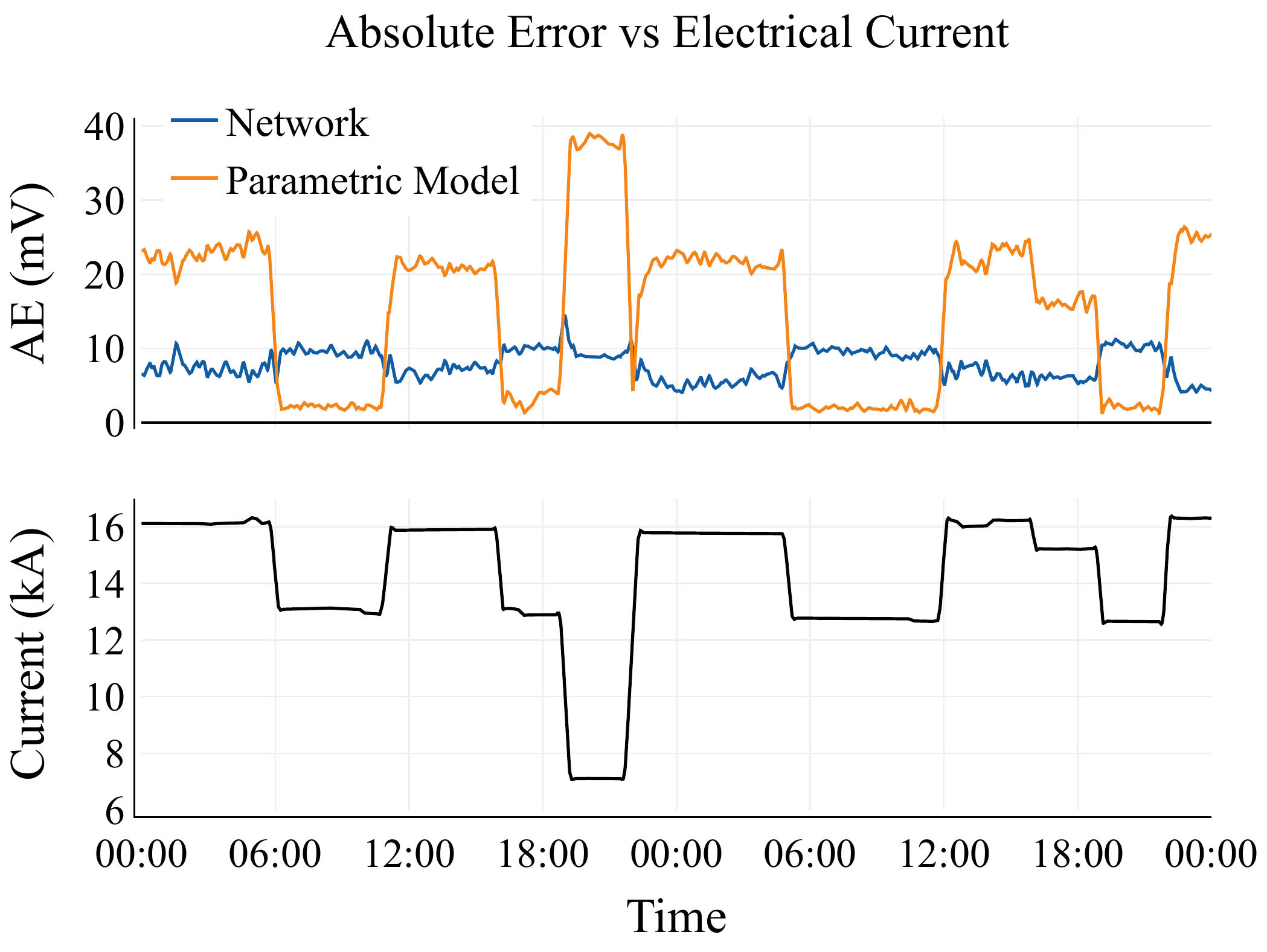}}

		\end{center}
		\vskip -0.2in
\end{figure}

In order to better visualize the prediction error, a closer look into a two-day period of the operation phase of cycle A is provided in Figure~\ref{fig: Models vs Electrical Current}.
We compare the behavior of the neural network against the parametric model and show that the voltage predicted by the network is more stable than the one predicted by the parametric model.
Indeed, the parametric model produces an unstable behavior because the experts' understanding of the reaction's kinetics is limited, and because, as previously explained, the parametric model must be simple in order to be trained using only the startup data. 

The response of the cell varies between a \emph{high load} and a \emph{low load}.
However, the parametric model is not capable of adapting its response to both of them, so it fits a response based on a \emph{medium load}.
Hence, an increased error is noticeable in Figure~\ref{fig: Models vs Electrical Current} when the current is around 16 kA or around 7 kA.
Indeed, the parametric model works better around 13 kA, which corresponds to a medium load.
The neural network, however, has no problem predicting the response for the different load levels, as it is not constrained to follow a linear relationship between the inputs.
The fact that our model considers the temporality of the time series also contributes to decreasing the absolute error, as the kinetics of the cell are taken into account by the predictor subnetwork.
This is especially helpful during load changes, as the model produces smooth transitions.

\subsection{Accuracy}
\label{section: Accuracy}

We are interested in comparing the accuracy between cycles (\emph{inter-cycle}), as well as comparing the accuracy inside the cycle (\emph{intra-cycle}).
We seek accurate results in both cases.

\subsubsection{Inter-Cycle}
\label{section: Inter-Cycle}

For each cell and cycle, we calculate the average absolute error for all the observations.
We then calculate a set of statistics for that averaged error, which are presented in Table~\ref{table: Inter-Cycle Error}.
Measuring the accuracy between cycles is vital, as a model must work correctly for as many cells and cycles as possible.
In fact, a model that is consistent with its predictions is preferred over a model that makes excellent predictions for some cells but weak ones for the rest of them.
Even if the former has a higher error on average.

Based on the results presented in Table~\ref{table: Inter-Cycle Error}, we conclude that our neural network is more reliable than the parametric model.
Not only is the average error lower, but also the standard deviation, which indicates that the error is less dispersed between different cells and cycles.
The parametric model produces more outliers, as its two highest percentiles present a significant deviation compared to those of the network.

\begin{table}[b]
		\caption{Statistics of the average absolute error of both models across the different cells and cycles, presented for different percentiles.}
		\label{table: Inter-Cycle Error}
		\vskip 0.1in
		\begin{center}
		\begin{small}
		\begin{tabular}{lcc}

		\toprule
		Stats & Neural Network [$mV$] & Parametric Model [$mV$] \\
		\midrule
				
		$\mu$ & 11.977 & 25.668 \\
		$\sigma$ & 11.503 & 30.304 \\
		$P_{25\%}$ & 5.470 & 8.398 \\
		$P_{50\%}$ & 8.432 & 16.225 \\
		$P_{75\%}$ & 13.488 & 30.760 \\
		$P_{90\%}$ & 24.167 & 53.262 \\
		$P_{95\%}$ & 35.461 & 82.745 \\
		$P_{99\%}$ & 62.676 & 157.488 \\

		\bottomrule

		\end{tabular}
		\end{small}
		\end{center}
		\vskip 0.25in
\end{table}

\subsubsection{Intra-Cycle}
\label{section: Intra-Cycle}

For each combination of cell and cycle, we calculate a set of statistics that represents the distribution of the error and average them across all the 6400 combinations.
Table~\ref{table: Intra-Cycle Error} presents these results.
The neural network obtains better results than the parametric model for all the statistics.
As expected, the average voltage is the same as in Table~\ref{table: Inter-Cycle Error}.

We can conclude that the accuracy intra-cycle is better than the inter-cycle one.
One possible explanation for this difference comes from the startup phase.
For the neural network, when the encoder works correctly, the error stays constant inside the cycle.
However, when the encoder fails to identify the degradation of the cell correctly, the accuracy of the whole cycle is penalized.
The same holds true for the parametric model.
When the startup is short, it does not have enough data to train, and the predictions are inaccurate for the whole cycle.

\begin{table}[t]
		\caption{Average statistics of the absolute error inside a cycle, presented for different percentiles.}
		\label{table: Intra-Cycle Error}
		\vskip 0.1in
		\begin{center}
		\begin{small}
		\begin{tabular}{lcc}
	
		\toprule
		Stats & Neural Network [$mV$] & Parametric Model [$mV$] \\
		\midrule
		
		$\mu$ & 11.977 & 25.668 \\
		$\sigma$ & 4.586 & 6.109 \\
		$P_{25\%}$ & 8.72 & 21.637 \\
		$P_{50\%}$ & 11.85 & 25.435 \\
		$P_{75\%}$ & 14.975 & 29.816 \\
		$P_{90\%}$ & 17.712 & 33.184 \\
		$P_{95\%}$ & 19.338 & 34.943 \\
		$P_{99\%}$ & 22.213 & 39.088 \\

		\bottomrule

		\end{tabular}
		\end{small}
		\end{center}
		\vskip -0.2in
\end{table}

\subsection{Fault Detection}
\label{section: Fault Detection}

As mentioned in the introduction, the end goal is to detect faults in cells before they happen.
The indicator is the divergence between the cell's predicted voltage by a trained model and the cell's measured voltage.

\begin{figure}[t]
		\caption{Divergence between the predicted voltage and the measured voltage during the 48 hours preceding a cell's fault.}
		\label{fig: Fault Detection}

		\vskip 0.2in
		\begin{center}
		\centerline{\includegraphics[width=\columnwidth]{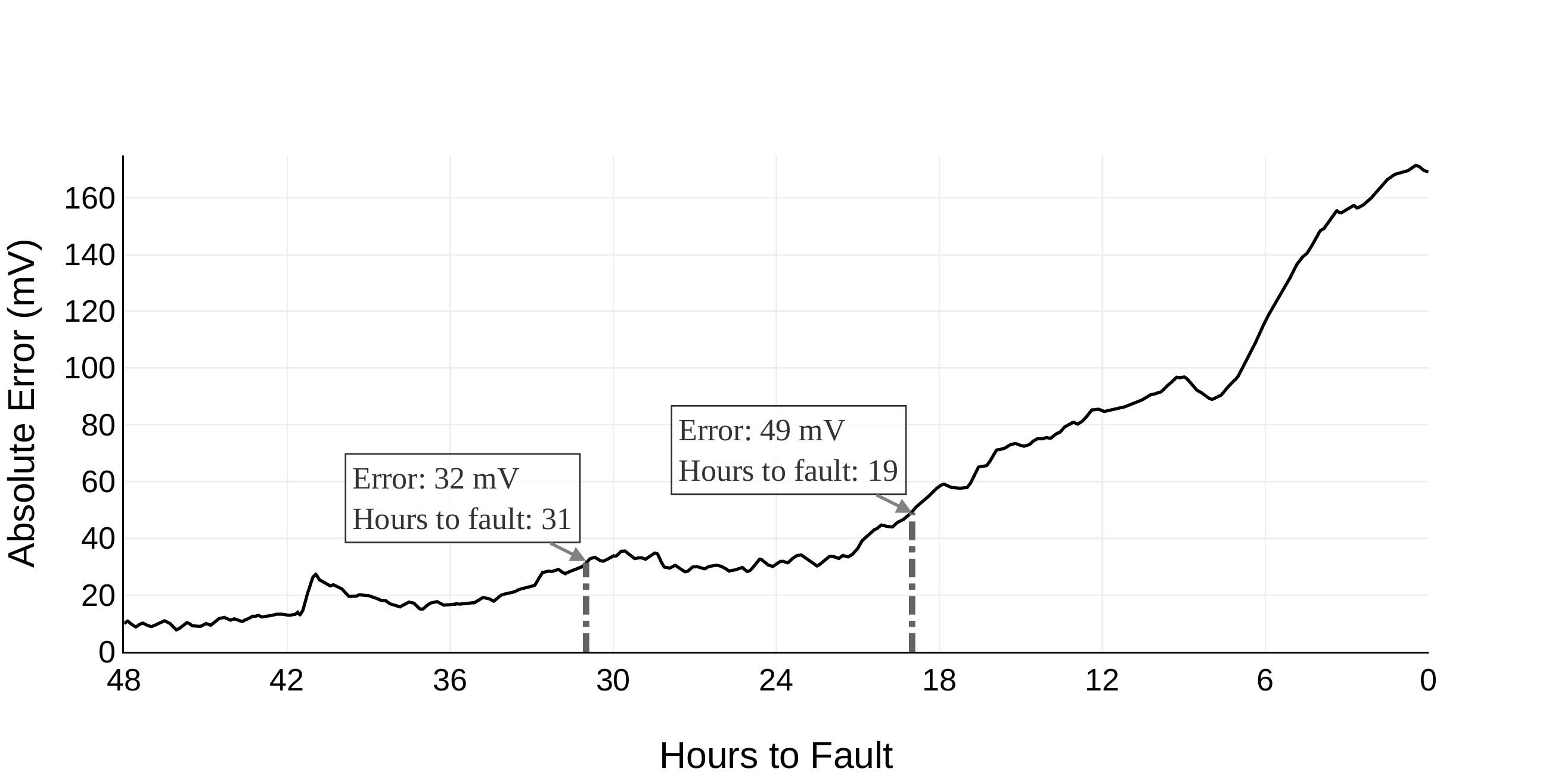}}

		\end{center}
		\vskip -0.2in
\end{figure}

In order to demonstrate that our neural network is appropriate for this task, we used the sequence of a faulty cell that was previously identified by an expert.
Figure~\ref{fig: Fault Detection} shows how the error between the predicted voltage and the measured voltage increases during the 48 hours that precede the fault.
The error increases slowly until it reaches a plateau.
It then stays stable for some hours until it starts to increase again with a more pronounced slope.
It is desired to detect the fault before the plateau, as that would give the plant's operators more time to react and plan the maintenance.
Therefore, the more accurate the model is, the earlier the impending fault can be detected.

We set the fault detection threshold to the average 99\textsuperscript{th} percentile intra-cycle error, as presented in the last row of Table~\ref{table: Intra-Cycle Error}.
In order to minimize the appearance of possible false positives, we add an extra tolerance of 10 mV.
This results in a threshold of 32 mV for the neural network and 49 mV for the parametric model.
In this case, the network was able to detect the fault 31 hours before it happened, which is a gain of 12 hours compared to the parametric model.

\section{Conclusion}
\label{section: Conclusion}

In this article, we present a new approach for detecting faults in electrochemical cells by using a neural network model composed of an encoder and a decoder.
Results show that this approach presents many advantages over expert-defined parametric models currently used for this task.
Namely, the key points of our approach are:

\begin{enumerate}

		\item \underline{Better accuracy}.
				Our model is capable of predicting the cell's voltage more accurately.
				We use a neural network that is not based on suppositions of the underlying chemical function.
				Instead, it is trained with a substantial amount of data coming from previous electrolyzers.
				It also takes into account the temporal relations that exist between the observations, thanks to its LSTM layers.
		
		\item \underline{Less human intervention required}.
				There is no need for an expert to spend time finding the specific parameters needed by each plant.
				This entails that the model may be deployed to more plants with little effort just by retraining the model, which in turn helps to improve their operational safety.
		
		\item \underline{Interpretability}.
				The output of the encoder can be plotted and explained, which helps the user to verify the correct functioning of the neural network.
				As the safety of the plant and its operators rely on the model, interpretability is very important.

		\item \underline{Continuously improving method}.
				As more data is collected from different cells and cycles, it is possible to retrain the model to include them.
				This will help reduce edge cases and the overall error of the neural network.

		\item \underline{Simple deployment}.
				A single model is valid for multiple cells and cycles without needing retraining once deployed.
				Therefore, the implementation of the model in the plant does not require extensive computing power.

\end{enumerate}

Thanks to this model, we can confidently say that we are one step closer to a fully automatic management of plant maintenance.

\section{Future Work}
\label{section: Future Work}

There are, however, some caveats and suggestions that could be further investigated in future models:

\begin{itemize}

		\item If the encoder does not find a good representation of a cell's startup, the voltage prediction for its operation phase will be incorrect.
				This happens scarcely, and, with more data available, it should occur even less often.
				Nevertheless, a safety fallback must be designed for these cases.

		\item It would be interesting to try to replace the LSTM encoder by an attention-based encoder in order to see if better accuracy is attained \cite{Vaswani2017}.

		\item Neural network's compression is an active field of research that could help in requiring less computing power and memory for making predictions \cite{Polino2018}.

\end{itemize}

% This section should not be numbered
\section{Acknowledgements}
\label{section: Acknowledgements}

We want to thank Mitacs and R2 Inc. for providing funding for this research.

% Number figures sequentially, placing the figure number and caption
% \emph{after} the graphics, with at least 0.1~inches of space before
% the caption and 0.1~inches after it, as in
% Figure~\ref{icml-historical}. The figure caption should be set in
% 9~point type and centered unless it runs two or more lines, in which
% case it should be flush left. You may float figures to the top or
% bottom of a column, and you may set wide figures across both columns
% (use the environment \texttt{figure*} in \LaTeX). Always place
% two-column figures at the top or bottom of the page.

% Note use of \abovespace and \belowspace to get reasonable spacing
% above and below tabular lines.

% Please use APA reference format regardless of your formatter
% or word processor. If you rely on the \LaTeX\/ bibliographic
% facility, use \texttt{natbib.sty} and \texttt{icml2020.bst}
% included in the style-file package to obtain this format.
\bibliography{article}
\bibliographystyle{article}

\end{document}